\documentclass{article}

\usepackage{nl_search}
\usepackage[utf8]{inputenc}
\usepackage[T1]{fontenc}    
\usepackage{hyperref}
\usepackage{url}         
\usepackage{booktabs}       
\usepackage{amsfonts}       
\usepackage{nicefrac}      
\usepackage{microtype}      
\usepackage{lipsum}
\usepackage{fancyhdr}      
\usepackage{graphicx}       
\usepackage{float}
\usepackage{enumitem}

\usepackage[title]{appendix}
\usepackage{authblk}

\renewcommand\Affilfont{\vspace{0.3em}\small}

\title{Leveraging LLMs to Enable Natural Language Search on Go-to-market Platforms}
\author[1]{Jesse Yao}
\author[2]{Saurav Acharya}
\author[2]{Priyaranjan Parida}
\author[2]{Srinivas Attipalli}
\author[2]{Ali Dasdan}
\affil[1]{University of California, Berkeley\thanks{Research performed
during a summer internship at Zoominfo in 2024.}}
\affil[2]{\small Zoominfo Technologies Inc.}
\affil[ ]{jessetyao@berkeley.edu, \{saurav.acharya, pri.parida,
  srinivas.attipalli, ali.dasdan\}@zoominfo.com}

\begin{document}
\maketitle
\vspace{-3.5em}
  
\begin{abstract}
Enterprise searches require users to have complex knowledge of
queries, configurations, and metadata, rendering it difficult for them
to access information as needed. Most go-to-market (GTM) platforms
utilize advanced search, an interface that enables users to filter
queries by various fields using categories or keywords, which,
historically, however, has proven to be exceedingly cumbersome, as
users are faced with seemingly hundreds of options, fields, and
buttons. Consequently, querying with natural language has long been
ideal, a notion further empowered by Large Language Models (LLMs).

In this paper, we implement and evaluate a solution for the Zoominfo
product for sellers, which prompts the LLM with natural language,
producing search fields through entity extraction that are then
converted into a search query. The intermediary search fields offer
numerous advantages for each query, including the elimination of
syntax errors, simpler ground truths, and an intuitive format for the
LLM to interpret.

We paired this pipeline with many advanced prompt engineering
strategies, featuring an intricate system message, few-shot prompting,
chain-of-thought (CoT) reasoning, and execution
refinement. Furthermore, we manually created the ground truth for 500+
natural language queries, enabling the supervised fine-tuning of
Llama-3-8B-Instruct and the introduction of sophisticated numerical
metrics.

Comprehensive experiments with closed, open source, and fine-tuned LLM
models were conducted through exact, Jaccard, cosine, and semantic
similarity on individual search entities to demonstrate the efficacy
of our approach. Overall, the most accurate closed model had an
average accuracy of 97\% per query, with only one field performing
under 90\%, with comparable results observed from the fine-tuned
models.
\end{abstract}

\keywords{Large Language Models (LLMs) \and Natural Language
  Processing (NLP) \and Prompt Engineering \and Fine-tuning \and
  Enterprise Search \and Go-to-market (GTM) Platforms}

\section{Introduction}

The ability to construct and execute enterprise searches is integral
for all sales and marketing departments in today’s data-driven world,
where quick access to relevant information is the key to
decision-making. This fundamental capability for many customers is
traditionally gated behind specialized understanding of queries,
configurations, and metadata, posing as a significant barrier to
widespread accessibility for non-technical users on go-to-market (GTM)
platforms, which help sellers and marketers with data and products to
acquire new customers and retain their existing customers.

Currently, enterprise searches on GTM platforms utilize a familiar
concept known as ``advanced search''. By entering keywords and
selecting categories in numerous fields, the user can filter based on
desired requirements and search terms, allowing for information
retrieval without the technical aspects. However, this interface often
presents users with a complex array of options, fields, and buttons,
leading to usability challenges and an overall arduous experience,
especially for larger databases~\cite{inproceedings,
  wilson2009evaluating}.

For instance, in the Zoominfo\footnote{Zoominfo is a leading provider
of the go-to-market data and products for sellers and marketers,
including ZI Sales.} product for sellers (called ZI Sales), company
attributes, one of dozens of fields, has over 200 options to choose
from. As a result, the concept of querying with natural language has
arisen as an ideal approach to bridge the gap between users and
databases, a notion further empowered by the emergence of Large
Language Models (LLMs).

This paper proposes an effective solution tailored to ZI Sales that
leverages LLMs to enable natural language queries, generating a
structured JSON file of search fields that is subsequently transformed
into Zoominfo’s query, known as a search service query. 

Generating the search service query directly from the LLM would likely
result in problems such as missing semicolons, misplaced fields, and
incorrect structure, making execution refinement impractical. Errors
would only be detected upon running the query, which is far too late
in the process. Consequently, the intermediary JSON configuration
provides several significant benefits for each query, with the primary
being the elimination of syntax errors. With proper prompt
engineering, the LLM has the ability to generate JSON files perfectly
due to the simplicity of the structure. Moreover, it is
straightforward to detect if the output is not a valid JSON file; in
such cases, we can prompt the LLM to regenerate the response with a
slightly altered prompt until a valid JSON file is produced. Another
overlooked benefit is the ease of creating the ground truth, allowing
the generation of an extensive ground truth, which substantially
enhances the accuracy and enables both numerical metrics and
supervised fine-tuning.

Each advanced search parameter was represented as a field in the JSON
file, each field being carefully selected to balance system accuracy
and the range of possible searches. Additionally, all fields were
classified into three categories: integer (e.g. revenue bounds),
categorical (e.g. US states), and free-text (e.g. titles); each field
type was evaluated on different metrics.

In addition to execution refinement, this reliable format was
complemented with several sophisticated prompt engineering strategies
including system messages, few-shot prompting, and chain-of-thought
(CoT) reasoning. We observed that explicitly detailing every aspect,
structure, and definition worked in our favor and significantly
increased our accuracy. Consequently, we meticulously defined every
field within the system message, resulting in a text of 3095
words. This was paired with few shot prompting and CoT to unveil
hidden logic and capture the ambiguities of the human language further
improving the precision of the model.

Each metric was applied on individual fields, with each type of field
evaluated on different metrics. Exact match was used for all fields,
Jaccard similarity was applied to categorical fields, and both cosine
and semantic similarity were tested against free-text fields. The
overall query average score was calculated as the average of all
field-specific metrics.

The most accurate and advanced model tested was Anthropic’s Claude 3.5
Sonnet~\cite{Sonnet-3.5}, which showcased a remarkable average query
accuracy of 97\%. It had only three fields with metrics below 95\% and
just one field under 90\%. However, due to the high cost and slower
nature of Sonnet 3.5, we also evaluated smaller LLMs to determine if
comparable results were achievable. Two models produced similar
results: Anthropic’s Claude 3 Haiku~\cite{haiku} and a fine-tuned
version of Llama3-8B-instruct~\cite{llama3}, which was trained with a
72\% training, 18\% validation, and 10\% testing split on Nvidia L4
GPUs~\cite{nvidia}. Both Haiku and fine-tuned Llama3 excelled in
several metrics compared to Sonnet 3.5; however, they fell slightly
short in the overall average performance. The differences between both
models and Sonnet 3.5 were primarily due to two fields: for Haiku,
company keywords and company attributes had deltas of around 4\% and
5\% respectively, while for Llama3, departments and titles had deltas
of 4\% and 9\% respectively.

Overall, this paper offers a natural language solution for enterprise
searches on GTM platforms. We show not only the feasibility but also
the extraordinary accuracy of such a solution, overshadowing the
original cumbersome advanced search. Although the translation is
specialized to this product, we believe the techniques we propose have
wider applicability.

\section{Related Work}
\label{sec:headings}

Translating from natural language to various query languages or query
interfaces is not new. We give a brief overview of the related work.

\subsection{NLP to SQL}

Natural language (NL) processing (NLP) to Structured Query Language
(SQL) is a well-known challenge within the machine learning industry,
with countless attempts made by companies even before the existence of
LLMs~\cite{Baig_Imran_Yasin_Butt_Muhammad,10.1145/3133887,
  xu2017sqlnetgeneratingstructuredqueries, 10.5555/1864519.1864543,
  cai2018encoderdecoderframeworktranslatingnatural,
  yu2018typesqlknowledgebasedtypeawareneural,
  yu2019spiderlargescalehumanlabeleddataset,
  zhong2017seq2sqlgeneratingstructuredqueries, Kim2020NaturalLT}. The
rise of LLMs has significantly increased the potential to address this
problem~\cite{Li_Boyan}, as demonstrated by various
efforts~\cite{li2023llmservedatabaseinterface,
  fan2024metasqlgeneratethenrankframeworknatural,ChRa2024,pourreza2024chasesqlmultipathreasoningpreference}.
However, despite these advancements, this idea has been especially
difficult to perfect with current models. For instance, on the BIRD
dataset, a collection of 12,751 NL and SQL pairs, even top industry
models like GPT-4~\cite{openai2024gpt4technicalreport} achieved only
54.89\% accuracy, significantly lower than the human benchmark of
92.96\%~\cite{li2023llmservedatabaseinterface}. On the same dataset,
the CHASE-SQL framework achieves a better accuracy of around
73\%~\cite{pourreza2024chasesqlmultipathreasoningpreference} but is
still lower than the human level.

Although our solution transforms natural language into a query, NLP to
SQL is fundamentally a different problem. SQL is designed for querying
raw data, which could potentially contain hundreds of tables, schemas,
and metadata. However, consumers should never have to query raw data
directly or understand complicated concepts like joining tables, case
statements, and aggregation. Consequently, enterprise searches on GTM
platforms provide a layer of abstraction, allowing consumers to
interact with a clean and unified dataset rather than a complex and
overwhelming collection of raw data. In essence, the data is already
preprocessed, so from the consumer standpoint, it appears as a single
table that can be queried with no technical knowledge, whether that be
through advanced search or natural language.

\subsection{LLM Based IR}

Information Retrieval (IR) systems, including search engines and
enterprise searches, have similarly been revolutionized by LLMs
\cite{zhu2024largelanguagemodelsinformation, e_search,
  li2024matchinggenerationsurveygenerative, Huang_2024,
  10.1093/jamia/ocae014,
  Structured_information_extraction}. Specifically for enterprise
searches, Bulfamante~\cite{e_search} proposed a pipeline that
leveraged semantic embeddings to identify and return documents most
relevant to the user's natural language query. Once retrieved, these
documents are processed by the LLM to craft detailed responses to the
user’s question. This retrieval augmented generation (RAG) system was
tested against RAGAS metrics and achieved commendable
results. Regarding entity extraction with LLMs, Dagdelen et
al.~\cite{Structured_information_extraction} explores its application
in chemistry, including ideas such as materials information
extraction, identifying chemical formulae, and recognizing inorganic
materials.

\begin{figure}
  \centering
  \includegraphics[width=0.75\textwidth]{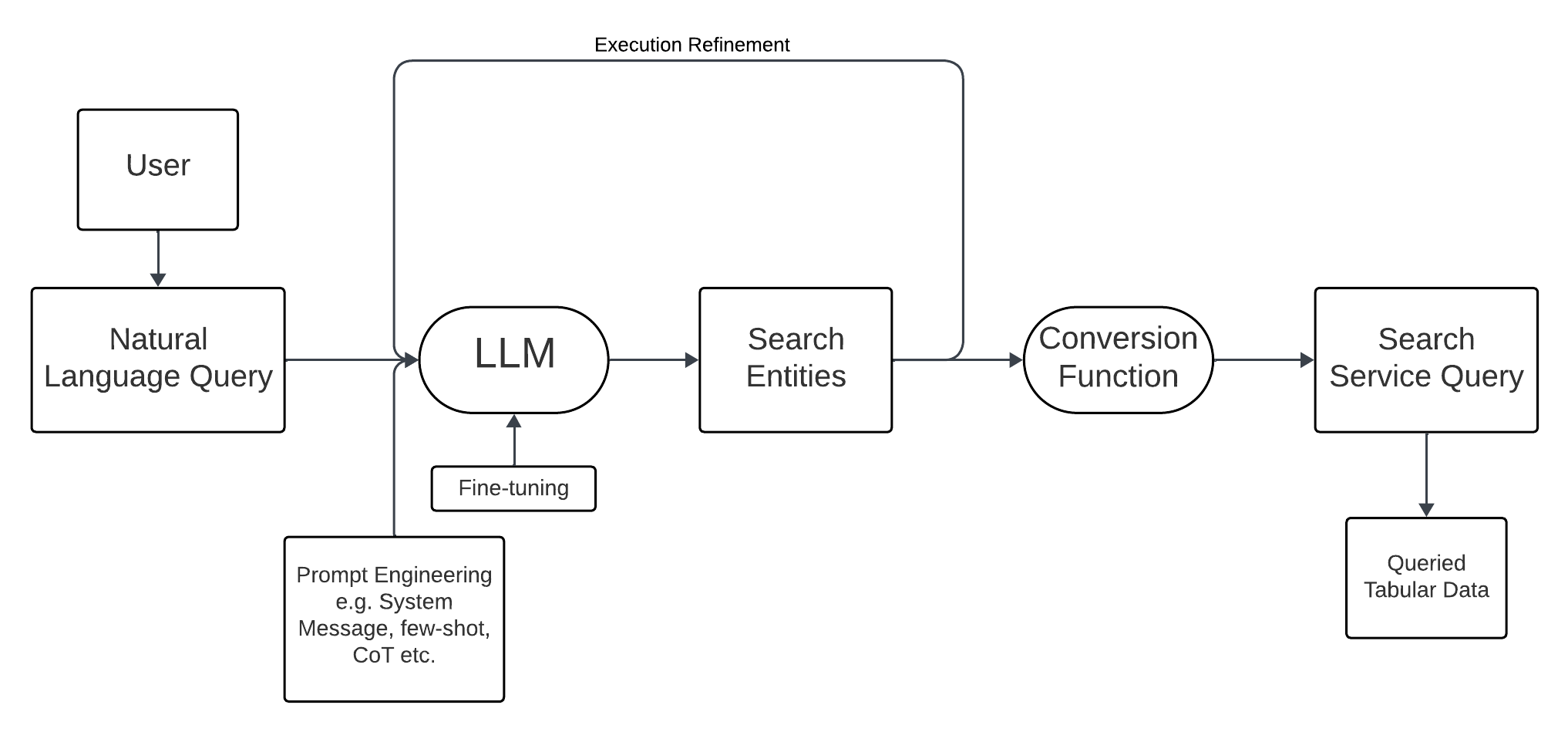}
  \caption{The system design of the search service. The user and user
    query are on the left side while the search service on the
    right. The LLM-based conversion with feedback is shown in the
    middle.}
  \label{fig:fig1}
\end{figure}

\section{Proposed Approach}

The proposed system in Fig.~\ref{fig:fig1} can be divided into two
phases: the conversion of a natural language query to search entities
and the mapping of search entities to a search service query. Before
presenting these phases, we give a brief introduction to the ZI Sales
product.

\subsection{ZI Sales: Zoominfo's sales product}

Zoominfo's sales product (called ZI Sales) is a search engine for
sellers to perform contact (business person) and company searches
using a free-form query as well as using various filters such as name,
title, revenue, employee count, and the like. The number of filters is
over 200.

\begin{figure}
  \centering
  \includegraphics[width=0.75\textwidth]{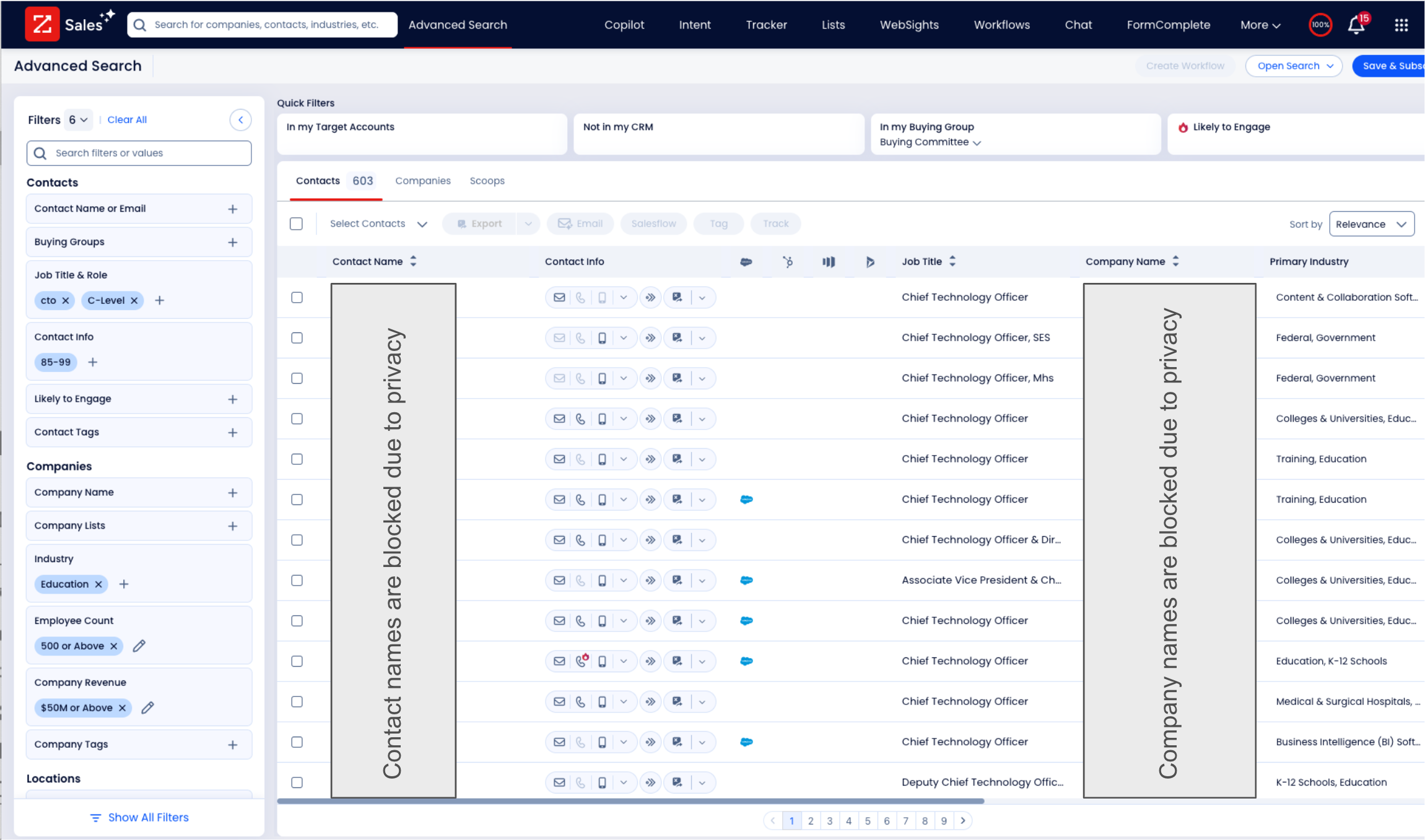}
  \caption{A screenshot of ZI Sales showing a search for the chief
    technology officers (CTOs) of companies in the education industry
    with at least 500 employees and at least \$50M revenue. The list
    has 603 names spread over multiple search results pages. The
    contact (person) and company names are not shown here due to
    privacy reasons.}
  \label{fig:zi}
\end{figure}

Fig.~\ref{fig:zi} shows a screenshot of the product showing the list
of the chief technology officers (CTOs) of companies in the education
industry with at least 500 employees and at least \$50M revenue. The
list has 603 names spread over multiple search results pages. The
``filters'' tab on the left of the screen shows many more filter
options. The person and company names are not shown in this figure due
to privacy reasons.

The sales product is a typical three-tier architecture with a search
backend layer, an application layer for the business logic, and a
presentation layer for inputting the query and filter interface and
presenting search results. The search backend provides join support on
top of an Apache Solr based search engine~\cite{SaChDa2024}.

\subsection{Natural Language to Search Entities}

First, a LLM is utilized to transform the natural language input into
a JSON output of advanced search fields. The intermediary search
entities provide several benefits, the most notable being the
elimination of syntax errors. Directly generating the search service
query from the LLM would result in issues such as missing semicolons,
misplaced fields, and incorrect structure, making execution refinement
impossible. Errors would only be detected upon running the query,
which is simply too late in the process. In contrast, the LLM, with
proper prompt engineering, is capable of generating JSON outputs
without syntax errors due to the simplicity of the structure. In
addition, it is trivial to detect if the output is not a JSON; in such
cases, we can simply prompt the LLM to regenerate the response with a
slightly altered prompt until a valid output is produced, a process
known as execution refinement.

Out of the 509 natural language queries submitted in the beta test,
zero queries from all Anthropic and fine-tuned models after prompt
engineering had syntax errors. In GPT-3.5-Turbo~\cite{gpt-3.5}, two
responses outputted “I’m sorry, I didn't quite catch that” or a
similar sentiment that were fixed after the execution refinement.

Furthermore, the JSON file provides a subtle yet significant
advantage: the complication of creating the ground truth. For each
query, the ground truth is a simple JSON file, which is notably easier
to create from a human perspective than a search service query; for a
search service query, it is quite arduous to even create 50 queries
let alone 500. This intermediary JSON approach allowed us to develop
an extensive ground truth, which naturally strengthens the accuracy,
allows for numerical metrics, and enables supervised fine-tuning.

\subsection{Search Entities to Search Service Query}

The transformation of the JSON search entities into a search service
query was mapped using a conversion function that also detects and
removes certain errors, ensuring smooth query execution.

\begin{table}[h!]
  \centering
  \caption{All the fields that are currently supported by the LLM. \textbf{Integer fields} can solely be integer values. \textbf{Categorical fields} selectively accept text values from a predefined wordbank. \textbf{Free-text fields}, provided the correct input type, can support anything.
}

\begin{tabular}{|p{0.2\textwidth}|p{0.2\textwidth}|p{0.1\textwidth}|p{0.35\textwidth}|}
\hline
\textbf{Fields} & \textbf{Input Type} & \textbf{Field Type} & \textbf{Example} \\
\hline
Company keywords & List of strings & Free-text & \verb|["crm software", "hospital"]| \\
\hline
Company name & List of strings & Free-text & \verb|["Zoominfo", "Chorus"]| \\
\hline
Location & Dictionary of string lists & Categorical & \thinspace\verb|{"us_states": ["Ohio"]}| \footnotemark\\

\hline
Revenue bounds & Integer & Integer & 500000 \\
\hline
Employee bounds & Integer & Integer & 5000 \\
\hline
Technologies & List of strings & Categorical & \verb|["Salesforce", "Snowflake"]| \\
\hline
Company attributes & List of strings & Categorical & \verb|["B2B", "SaaS"]| \\
\hline
Company type & List of strings & Categorical & \verb|["Public", "Non-profit"]| \\
\hline
Company news & List of strings & Categorical & \verb|["Event", "Award"]| \\
\hline
Management levels & List of strings & Categorical & \verb|["C-Level", "VP-Level"]| \\
\hline
Departments & List of strings & Categorical & \verb|["Sales", "Legal"]| \\
\hline
Person name & String & Free-text & "Ali Dasdan" \\
\hline
Titles & List of strings & Free-text & \verb|["Data Scientist", "Lawyer"]| \\
\hline
Contact info & List of strings & Categorical & \verb|["Phone", "Email"]| \\
\hline
Person or company & String & Categorical & "person" \\
\hline
\end{tabular}
\label{tab:example}
\end{table}
\footnotetext{{All location fields include US States, CA provinces, US and CA Metros, and an others fields.}}

\section{Fields and Query Types}

An advanced search field is a designated parameter in a search
interface used to refine queries based on specific criteria, such as
categories, keywords, or metadata. For instance, the “company\_name”
field will exclusively accept company names, ensuring that only those
names are outputted in the final table. Listed in
Table~\ref{tab:example} are all the supported fields. These fields
were explicitly selected to strike a delicate balance since too many
fields may reduce the system’s accuracy, while too few may not
adequately cover the desired range of searches.

With respect to the desired range, fewer than 10 out of the 509
queries were supported by the Advanced search and not by the
LLM. Among them, only two were affected by a limited word bank in a
categorical field while the remainder involved searches based on
intent. In addition, approximately 100 queries were not supported by
the Advanced search itself, likely due to the unclear instructions in
the beta test. For instance, a query asking for account information
like “find me a list of C-level IT contacts in New York that I have
not exported to Salesforce” or ambiguous queries such as just entering
a period. Ultimately, around 400 queries remain for the experiments.

\section{LLM Optimization}

We discuss multiple optimization techniques we used in our research.

\subsection{Prompt Engineering}

Prompt Engineering is the practice of designing and refining input
prompts to optimize the performance of the LLM. Beyond JSON outputs
and execution refinement, other advanced prompt engineering techniques
were deployed within the system as well.

\subsubsection{System Message}

The system message defines the role of the LLM, the schema of the
table, and the hidden logic within the task. For our use case, there
is only one schema for the Zoominfo sales product, allowing the complete
integration of the schema into the system message; however, varying
schemas and fields would require the use of retrieval-augmented
generation (RAG) rather than a static message. Consequently, the
top-performing model, Claude 3.5 Sonnet, utilized a system message
that contained 3095 words.

In the system message, the schema of the Zoominfo sales product is
meticulously defined. For instance, company keywords were described as
follows:

\begin{verbatim}
"company_keywords": List[str]
Description: Filter for companies with these specialties, products, services, or 
industries in which they operate, etc. For example, "car manufacturer", "microwave 
ovens", "crm software", "cloud security", "pharmaceuticals", "hospital", "software", …
\end{verbatim}

For categorical fields, the entire word bank was provided if possible
(unless it was trivial, like the 50 states). We observed that
specifying every entry permitted by the field significantly increased
the accuracy of the particular field. For example, management levels
were described as follows:

\begin{verbatim}
"management_levels": List[str] 
Description: Filter for contacts that are people managers or individual contributors 
(Non-Managers), and their level of management responsibility (not technical level). 
Guidelines: Must be zero or more of these options. Leave this empty if "titles" is not 
empty. Do not guess the level based on specific requested titles. 
Choices (comma separated): 
C-Level,VP-Level,Director,Manager,Non-Manager
\end{verbatim}

Notably, all possible entries were predefined along with the logic of
mutual exclusivity between management\_levels and titles. However,
although defined in the field, simply stating the logic was not
sufficiently effective for the LLM to understand, necessitating
further prompt engineering.

\subsubsection{Few-Shot Learning}

Few-shot learning involves providing the LLM with a few examples
(shots) to help the model understand the desired output format and
reasoning. In the context of our model, it entails presenting the
model with question-answer pairs, where the question is a natural
language query and the answer takes the form of a JSON file. One shot
would resemble the following:

\begin{verbatim}
Prompt: decision makers at Zoominfo and Chorus in the us

Answer in json format:
{
 "company_name": ["Zoominfo", "Chorus"],
 "management_levels": ["C-Level","VP-Level"],
 "location": {
   "us_states": ["United States"]
 },
 "person_or_company": "person"
}
\end{verbatim}

Claude 3.5 Sonnet used a total of 11 shots. Each shot was selectively
chosen to unveil hidden logic and clarify the ambiguity of the human
language. For example, what does the term “decision makers” mean?  We
have defined it to the model as C-Level and VP-Level, but this is open
to interpretation. These rationales were explicitly stated by another
technique known as chain-of-thought (CoT) prompting.

\subsubsection{Chain-of-Thought (CoT)}

CoT prompting is defined as guiding the LLM step-by-step through the
exact reasoning process to determine the desired output. In our
context, it serves to precisely describe and justify the expected JSON
from the natural language query. Since natural language is potentially
quite vague, or logic that seems intuitive to a human might not be
intuitive to a LLM, CoT aids the LLM by providing explicit
rationale. For instance, in the example above:

\begin{verbatim}
Reasoning:
Notice that "decision makers" refers to C-Level and VP-Level executives.

Also notice that because titles and management_levels are mutually exclusive and since 
a specific title was not specified, the management_levels field can exist. Note that 
titles and departments are also mutually exclusive, but management_levels and departments 
are not.

Notice that because two companies were mentioned, company_name has two entries in the list
\end{verbatim}

With this one example, three pieces of logic were defined for the
LLM. We further noticed that severe specificity often worked in our
favor as it would phase out certain careless errors, such as providing
a string for one company name as opposed to a list with one element.

\subsection{Fine-Tuning}

Fine-tuning models involves additional training with datasets to
improve the LLM’s performance on specific tasks. This process of
adjusting the model’s parameters allows the model to adapt its general
knowledge to a more nuanced target domain, resulting in more accurate
and relevant outputs. Fine-tuning can be further enhanced by Low-Rank
Adaptation (LoRA)~\cite{hu2021loralowrankadaptationlarge}, which
involves the decomposition of lower weight matrices into
lower-dimensional representations. LoRA significantly reduces the
computation resources required and shortens the training time, leading
to a more efficient training.

We conducted supervised fine-tuning on the open source model
Llama3-8B-Instruct, using the ground truth as a labeled dataset (note
that Llama3-70B was not tested due to limited GPU resources). After
setting aside 10\% of the data for the test set, the model was trained
using an 80\% training and 20\% validation split and computed on
Nvidia L4 GPUs~\cite{nvidia} through Google Cloud.

\section{Experiments}

We discuss the similarity metrics, experiments, and results.

\subsection{Similarity Metrics}

To measure the quality of the search results at a granular level, we
implemented similarity metrics specific to each field listed in
Table~\ref{tab:example}, with the overall query accuracy as the
average across every field. We utilized four different metrics, each
tailored to integer, categorical, and free-text fields, and compared
them against the ground truth. The four metrics are as follows:

\begin{itemize}[noitemsep, left=0pt]
    \item \textit{Exact match} - score of 1 if they exactly match and
      0 otherwise. To remain consistent, all strings were made
      lowercase and all lists were sorted before validating an exact
      match.
    \item \textit{Jaccard similarity} - the total number of matches
      divided by the total number of unique elements between two sets.
    \item \textit{Cosine similarity} - the similarity between two
      texts based on the cosine of the angle between their frequency
      vector representations.
    \item \textit{Semantic similarity} - the similarity between the
      meaning of two text segments by utilizing word
      embeddings~\cite{Chandrasekaran_2021}.
\end{itemize}

Exact match was applied to all fields, Jaccard was used for
categorical fields, and cosine and semantic similarity were utilized
for free-text fields. In cases where a field was correctly omitted,
all metrics of that field would receive a score of one and a zero if
the field was extra or missing.

\subsection{Evaluation}

In the evaluation of prompt engineering techniques, fine-tuning, and
LLM models, we divided the experiment into two parts: closed models
and open source models.

\subsubsection{Closed Models}

Out of Sonnet 3.5, Haiku, and GPT-3.5-Turbo, the most accurate model,
after prompt engineering and execution refinement, was Sonnet 3.5,
which demonstrated an average query accuracy of 97\% across all
metrics. Notably, in this calculation, free text fields were weighed
three times more than integer fields, as they were assessed using
three distinct metrics compared to a singular one. However, since free
text fields are the worst-performing, the 97\% actually represents a
lower and more conservative estimate. Due to the imbalance of weights,
it is imperative to examine performance at the individual field level,
where results are equally, if not more, impressive than the overall
query score. Specifically, across all individual fields, only three
fields in Sonnet 3.5 exhibit performance statistics below 95\%, with
merely one field falling below the 90\% threshold.

\begin{figure}
  \centering
  \includegraphics[width=1\textwidth]{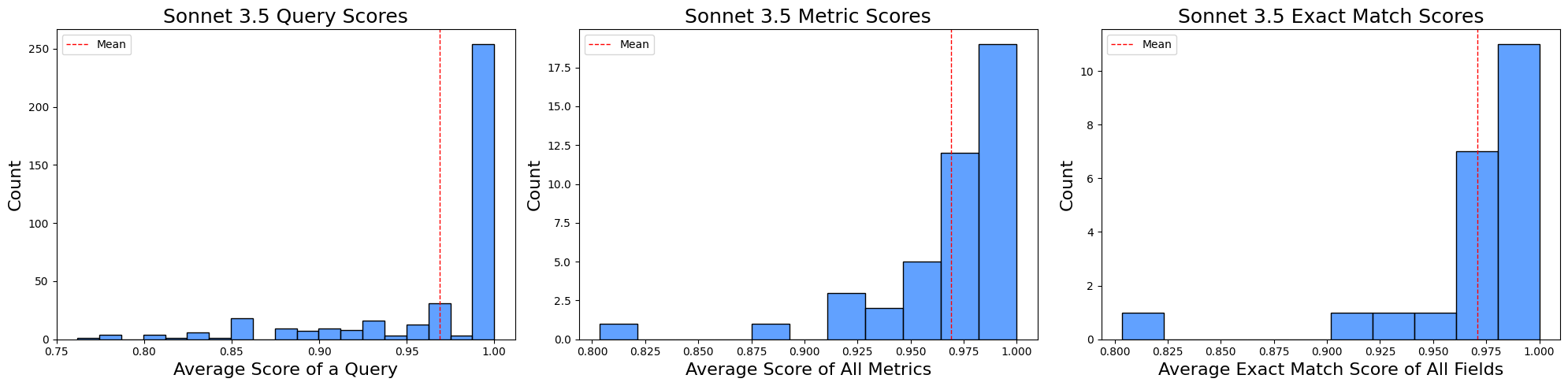}
  \caption{Score histograms and average (mean) scores for Sonnet
    3.5. From left to right: the average score of every metric for an
    individual query; the average scores for all metrics; the score of
    every exact match metric.  }
  \label{fig:fig2}
\end{figure}

As shown in Fig.~\ref{fig:fig2}, most fields score relatively well
even for exact match metrics. The three lowest-performing exact match
fields are, unsurprisingly, all free text fields, which, as
illustrated in Table~\ref{tab:table2}, perform better with regard to cosine and
semantic similarity. Intriguingly, all categorical text fields
performed exceptionally well, with no metric below 95\%, likely due to
the rigorous pre-definition of word banks. Even subtle keywords, such
as “northeast,” referring to a collection of nine US states, were
captured with reliable accuracy. For the integer fields-specifically
the exact matches for the revenue and employee bounds-the scores were
0.997, 1.0, 1.0, and 0.995, respectively ordered as revenue (lower, upper)
followed by employee (lower, upper).

\begin{table}
  \centering
 \caption{Average similarity scores for free text fields for Sonnet
   3.5.}
  \begin{tabular}{llll}
    \toprule
    \cmidrule(r){1-2}
Field & Exact Match & Cosine & Semantic \\
    \midrule
Company Keywords & 0.804 & 0.890 & 0.918 \\
Company Name & 0.920 & 0.928 & 0.930 \\
Person Name & 0.992 & 0.994 & 0.993\\
Titles & 0.941 & 0.954 & 0.959 \\
    \bottomrule
  \end{tabular}
  \label{tab:table2}
\end{table}

Sonnet 3.5, however, operates around three times slower (and is 12
times more expensive circa August 2024) than Haiku, Anthropic’s most
affordable model. This raises the question: Can we achieve
sufficiently satisfactory results with a cheaper model?

\begin{figure}
  \centering
  \includegraphics[width=1\textwidth]{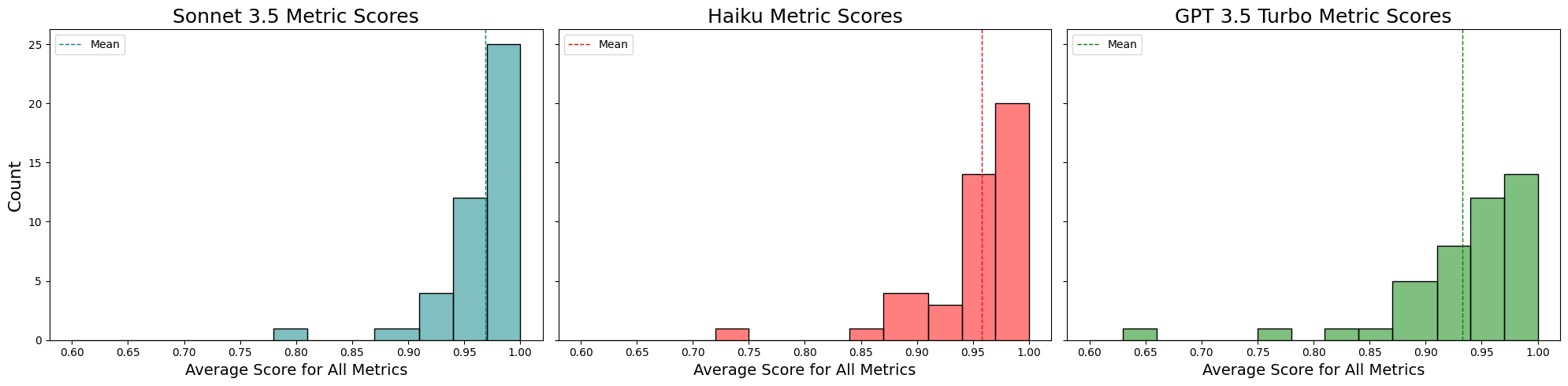}
  \caption{The score histograms and average (mean) scores of Sonnet
    3.5 vs Haiku vs GPT-3.5-Turbo per metric. Note that each LLM has
    unique system messages and sets of shots to optimize its own
    score.}
  \label{fig:fig3}
\end{figure}

Evidently from Fig.~\ref{fig:fig3}, GPT-3.5-Turbo (average score of
0.933) is clearly outperformed by Sonnet 3.5. Haiku (average score of
0.958), in contrast, demonstrates a remarkably closer performance to
Sonnet 3.5 and even outperforms it in several metrics. In fact, the
distinction between Sonnet 3.5 and Haiku can be encapsulated by just
two fields, as illustrated in Table~\ref{tab:table3}.

\begin{table}
 \caption{Comparison of Haiku vs Sonnet 3.5}
  \centering
  \begin{tabular}{llllll}
    \toprule
    \textbf{Fields} & \textbf{Model} & \textbf{Exact} & \textbf{Jaccard} & \textbf{Cosine} & \textbf{Semantic} \\
    \midrule
    Company Keywords & Haiku & 0.744 & - & 0.846 & 0.872 \\
    \quad\quad\quad\quad\quad$\Rightarrow$ & Sonnet 3.5 & 0.804 & - & 0.890 & 0.918 \\
    Company Attributes & Haiku & 0.915 & 0.918 & - & - \\
    \quad\quad\quad\quad\quad$\Rightarrow$ & Sonnet 3.5 & 0.963 & 0.965  & - & -\\
    \bottomrule
  \end{tabular}
  \\ \leftskip=80pt (Note that some metrics are not applicable due to the field type)
  \label{tab:table3}
\end{table}

\subsubsection{Open Source Models}

The final model is a fine-tuned version of Llama3-8B-Instruct, an open
source model. Despite a measly 8 billion parameters, the fine-tuned
model (average score of 0.956) significantly surpasses both the base
Llama3 model and GPT-3.5-Turbo, performing comparably with the
Anthropic models. Similar to Haiku, it excels in several metrics
compared to Sonnet 3.5, although it falls slightly short in the
overall average performance.

Interestingly, the fine-tuned model outperforms Sonnet 3.5 in company
name recognition, achieving a score of 0.936 compared to 0.914, and
shows improved differentiation between US metro areas. However, the
slightly lower average for the fine-tuned model is primarily due to
two key fields: departments and titles. While Sonnet 3.5 scored 0.989
and 0.979 in these fields, respectively, the fine-tuned model only
achieved results of 0.945 and 0.893. This near 9\% delta in the titles
field highlights Sonnet’s superior accuracy among all the models
tested.

\begin{figure}[H]
  \centering
  \includegraphics[width=1\textwidth]{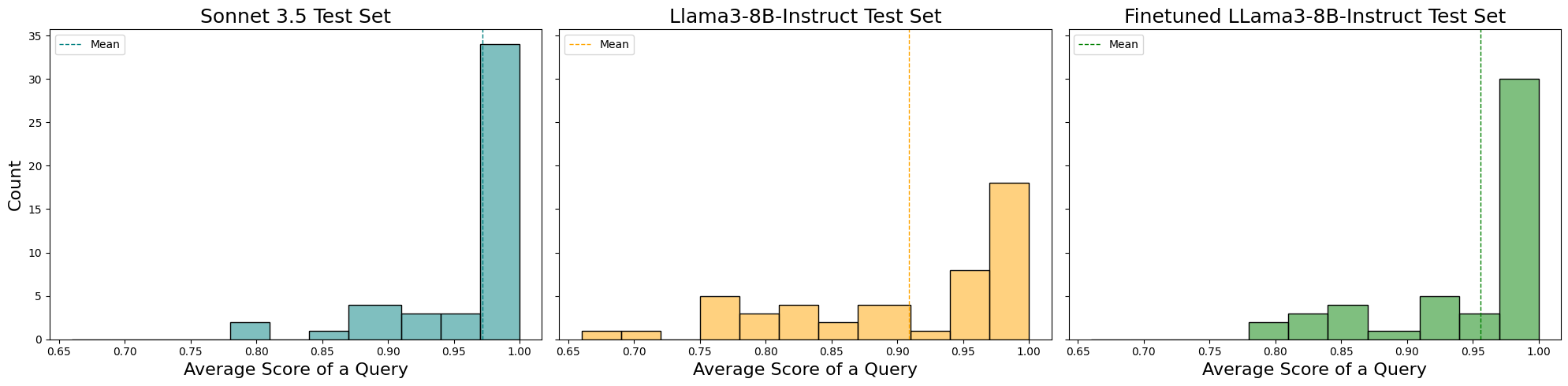}
  \caption{The score histograms and average (mean) scores for the
    prompts within the test set. Around 130 queries of synthetic data
    were also created to test the difference between the models.}
  \label{fig:fig4}
\end{figure}

\section{Conclusion}

The integration of natural language to execute enterprise searches
represents a significant advantage for users on GTM platforms to be
able to extract information. This study demonstrated the effectiveness
of querying an enterprise database using natural language inputs,
converting them into structured JSON queries via an intermediary
format. By leveraging advanced prompt engineering techniques such as
system messages, few-shot prompting, and chain-of-thought reasoning,
alongside rigorous evaluation against the ground truth and execution
refinement, this approach achieved exceptional results in accuracy for
every field supported. The evaluation of query performance used
sophisticated metrics including exact match, Jaccard, cosine, and
semantic similarity highlighted robust average accuracy rates of 97\%
across queries, with only three fields scoring below 95\% for Claude
3.5 Sonnet. Furthermore, fine-tuning Llama3-8B-Instruct yielded
similar results despite the disparity in the number of parameters and
the size of the model.

Our methodology solidifies not only the feasibility of a natural
language enterprise search but also the potential and accuracy of such
an approach. It offers an effective and streamlined alternative to
existing GTM platform's advanced search, and as technologies continue
to evolve, this concept of natural language will further enhance the
usability and efficiency of the information retrieval processes in the
future.

\section{Future Work}

Due to limitations on GPU hardware, we were constrained to only
fine-tuning the Llama3-8B-Instruct model. In the future, we plan to
fine-tune the latest Llama models, Llama3.1-70B and 405B
\cite{llama3.1}. Furthermore, we aim to implement advanced queries
with negation logic and support intent queries, while maintaining the
current accuracy.

\bibliographystyle{unsrt}  
\bibliography{references}

\begin{thebibliography}{10}

\bibitem{inproceedings}
R.~White and D.~Morris.
\newblock Investigating the querying and browsing behavior of advanced search
  engine users.
\newblock In {\em Proc. of the 30th Annual International ACM SIGIR Conf. on
  Research and Development in Information Retrieval}, pages 255--262, 07 2007.

\bibitem{wilson2009evaluating}
M.L. Wilson, M.C. Schraefel, and R.W. White.
\newblock Evaluating advanced search interfaces using established
  information-seeking models.
\newblock {\em J. of the American Society for Information Science and
  Technology (JASIST)}, 60(7):1407--1422, July 2009.

\bibitem{Sonnet-3.5}
Anthropic.
\newblock Introducing {C}laude 3.5 {S}onnet., 2024.

\bibitem{haiku}
Anthropic.
\newblock Claude 3 {H}aiku: Our fastest model yet, 2024.

\bibitem{llama3}
Meta.
\newblock Llama 3, 2024.

\bibitem{nvidia}
Nvidia.
\newblock Nvidia l4 {GPU} {A}ccelerator - product brief, 2023.

\bibitem{Baig_Imran_Yasin_Butt_Muhammad}
M.~Baig, A.~Imran, A.~Yasin, A.H. Butt, and M.I. Khan.
\newblock Natural language to {SQL} queries: A review.
\newblock {\em Int. J. of Innovations in Sci. and Tech.}, 4:147--162, 02 2022.

\bibitem{10.1145/3133887}
N.~Yaghmazadeh, Y.~Wang, I.~Dillig, and T.~Dillig.
\newblock {SQLizer}: Query synthesis from natural language.
\newblock {\em Proc. ACM Program. Lang.}, 1(OOPSLA), oct 2017.

\bibitem{xu2017sqlnetgeneratingstructuredqueries}
X.~Xu, C.~Liu, and D.~Song.
\newblock {SQLNet}: Generating structured queries from natural language without
  reinforcement learning.
\newblock Arxiv, 2017.

\bibitem{10.5555/1864519.1864543}
J.M. Zelle and R.J. Mooney.
\newblock Learning to parse database queries using inductive logic programming.
\newblock In {\em Proc. of the 13th National Conf. on Artificial Intelligence -
  V.2}, AAAI'96, page 1050–1055. AAAI Press, 1996.

\bibitem{cai2018encoderdecoderframeworktranslatingnatural}
R.~Cai, B.~Xu, X.~Yang, Z.~Zhang, Z.~Li, and Z.~Liang.
\newblock An encoder-decoder framework translating natural language to database
  queries.
\newblock Arxiv, 2018.

\bibitem{yu2018typesqlknowledgebasedtypeawareneural}
T.~Yu, Z.~Li, Z.~Zhang, R.~Zhang, and D.~Radev.
\newblock {TypeSQL}: Knowledge-based type-aware neural text-to-{SQL}
  generation.
\newblock Arxiv, 2018.

\bibitem{yu2019spiderlargescalehumanlabeleddataset}
T.~Yu, R.~Zhang, K.~Yang, M.~Yasunaga, D.~Wang, Z.~Li, J.~Ma, I.~Li, Q.~Yao,
  S.~Roman, Z.~Zhang, and D.~Radev.
\newblock Spider: A large-scale human-labeled dataset for complex and
  cross-domain semantic parsing and text-to-{SQL} task.
\newblock Arxiv, 2019.

\bibitem{zhong2017seq2sqlgeneratingstructuredqueries}
V.~Zhong, C.~Xiong, and R.~Socher.
\newblock {Seq2SQL}: Generating structured queries from natural language using
  reinforcement learning.
\newblock Arxiv, 2017.

\bibitem{Kim2020NaturalLT}
H.~Kim, B.-H. So, W.-S. Han, and H.~Lee.
\newblock Natural language to {SQL}: Where are we today?
\newblock {\em Proc. VLDB Endow.}, 13:1737--1750, 2020.

\bibitem{Li_Boyan}
B.~Li, Y.~Luo, C.~Chai, G.~Li, and N.~Tang.
\newblock The dawn of natural language to {SQL}: Are we fully ready?
\newblock Arxiv, 2024.

\bibitem{li2023llmservedatabaseinterface}
J.~Li, B.~Hui, G.~Qu, J.~Yang, B.~Li, B.~Li, B.~Wang, B.~Qin, R.~Cao, R.~Geng,
  N.~Huo, X.~Zhou, C.~Ma, G.~Li, K.C.C. Chang, F.~Huang, R.~Cheng, and Y.~Li.
\newblock Can {LLM} already serve as a database interface? a big bench for
  large-scale database grounded text-to-{SQL}s.
\newblock Arxiv, 2023.

\bibitem{fan2024metasqlgeneratethenrankframeworknatural}
Y.~Fan, Z.~He, T.~Ren, C.~Huang, Y.~Jing, K.~Zhang, and X.S. Wang.
\newblock Metasql: A generate-then-rank framework for natural language to {SQL}
  translation.
\newblock Arxiv, 2024.

\bibitem{ChRa2024}
A.~Chopra and R.~Azam.
\newblock Enhancing natural language query to {SQL} query generation through
  classification-based table selection.
\newblock In L.~Iliadis, I.~Maglogiannis, A.~Papaleonidas, E.~Pimenidis, and
  C.~Jayne, editors, {\em Engineering Applications of Neural Networks}, pages
  152--165, Cham, 2024. Springer Nature Switzerland.

\bibitem{pourreza2024chasesqlmultipathreasoningpreference}
M.~Pourreza, H.~Li, R.~Sun, Y.~Chung, S.~Talaei, G.T. Kakkar, Y.~Gan,
  A.~Saberi, F.~Ozcan, and S.O. Arik.
\newblock {CHASE-SQL}: Multi-path reasoning and preference optimized candidate
  selection in text-to-{SQL}.
\newblock Arxiv, 2024.

\bibitem{openai2024gpt4technicalreport}
{OpenAI}.
\newblock {GPT}-4 technical report.
\newblock Arxiv, 2024.

\bibitem{zhu2024largelanguagemodelsinformation}
Y.~Zhu, H.~Yuan, S.~Wang, J.~Liu, W.~Liu, C.~Deng, H.~Chen, Z.~Dou, and J.-R.
  Wen.
\newblock Large language models for information retrieval: A survey.
\newblock Arxiv, 2024.

\bibitem{e_search}
D.~Bulfamante.
\newblock Generative enterprise search with extensible knowledge base using
  {AI}.
\newblock Master's thesis, Politecnico di Torino, 2023.

\bibitem{li2024matchinggenerationsurveygenerative}
X.~Li, J.~Jin, Y.~Zhou, Y.~Zhang, P.~Zhang, Y.~Zhu, and Z.~Dou.
\newblock From matching to generation: A survey on generative information
  retrieval.
\newblock Arxiv, 2024.

\bibitem{Huang_2024}
Y.~Huang and J.X. Huang.
\newblock Exploring {ChatGPT} for next-generation information retrieval:
  Opportunities and challenges.
\newblock {\em Web Intelligence}, 22(1):31–44, March 2024.

\bibitem{10.1093/jamia/ocae014}
W.~Hersh.
\newblock Search still matters: information retrieval in the era of generative
  {AI}.
\newblock {\em J. of the American Medical Informatics Association}, page
  ocae014, 01 2024.

\bibitem{Structured_information_extraction}
J.~Dagdelen, A.~Dunn, S.~Lee, N.~Walker, A.S. Rosen, G.~Ceder, K.A. Persson,
  and A.~Jain.
\newblock Structured information extraction from scientific text with large
  language models.
\newblock {\em Nature Communications}, page 1418, 2024.

\bibitem{SaChDa2024}
H.~Sarkezians, J.~Chou, and A.~Dasdan.
\newblock How we built search for go-to-market platforms at {Z}oominfo.
\newblock Rockset Index Conference, 2024.

\bibitem{gpt-3.5}
{OpenAI}.
\newblock {GPT}-3.5 {T}urbo, 2023.

\bibitem{hu2021loralowrankadaptationlarge}
E.J. Hu, Y.~Shen, P.~Wallis, Z.~Allen-Zhu, Y.~Li, S.~Wang, L.~Wang, and
  W.~Chen.
\newblock {LoRA}: Low-rank adaptation of large language models.
\newblock Arxiv, 2021.

\bibitem{Chandrasekaran_2021}
D.~Chandrasekaran and V.~Mago.
\newblock Evolution of semantic similarity—a survey.
\newblock {\em ACM Computing Surveys}, 54(2):1–37, February 2021.

\bibitem{llama3.1}
Meta.
\newblock Introducing {L}lama 3.1: Our most capable models to date, 2024.

\end{thebibliography}

\end{document}